\documentclass{article}

\usepackage{arxiv}

\usepackage[utf8]{inputenc} 
\usepackage[T1]{fontenc}    
\usepackage{hyperref}       
\usepackage{url}            
\usepackage{booktabs}       
\usepackage{amsfonts}       
\usepackage{nicefrac}       
\usepackage{microtype}      
\usepackage{lipsum}		
\usepackage{graphicx}
\usepackage{natbib}
\usepackage{doi}

\usepackage{amsmath,amssymb}
\usepackage{multirow}
\usepackage{float}
\usepackage{makecell}
\usepackage{algorithm}
\usepackage{algorithmic}
\usepackage{bm}
\usepackage{booktabs}
\usepackage{array, caption, threeparttable}
\usepackage[section]{placeins}
\usepackage[font=small,labelfont=bf]{caption}

\title{Feature Extraction Framework based on Contrastive Learning with Adaptive Positive and Negative Samples}

\author{ {\hspace{1mm}Hongjie~Zhang}\\
	College of Information and Electrical Engineering\\
	China Agricultural University\\
	Beijing 100083, PR China \\
	\texttt{zhanghongjie@cau.edu.cn} }

\renewcommand{\shorttitle}{}

\begin{document}
\maketitle

\begin{abstract}
	In this study, we propose a feature extraction framework based on contrastive learning with adaptive positive and negative samples (CL-FEFA) that is suitable for unsupervised, supervised, and semi-supervised single-view feature extraction.
	CL-FEFA constructs adaptively the positive and negative samples from the results of feature extraction, which makes it more appropriate and accurate.  Thereafter, the discriminative features are re extracted to according to InfoNCE loss based on previous positive and negative samples, which will make the intra-class samples more compact and the inter-class samples more dispersed. At the same time, using the potential structure information of subspace samples to dynamically construct positive and negative samples can make our framework more robust to noisy data. Furthermore, CL-FEFA considers the mutual information between positive samples, that is, similar samples in potential structures, which provides theoretical support for its advantages in feature extraction. The final numerical experiments prove that the proposed framework has a strong advantage over the traditional feature extraction methods and contrastive learning methods.
\end{abstract}

\section{Introduction}
Currently, high-dimensional data is widely used in pattern recognition and data mining, which leads to high storage overhead, heavy computation, and excessive time consumption apart from causing the problem known as ``curse of dimensionality''. A significant way to address these issues is feature extraction, which transforms the original high-dimensional spatial data into a low-dimensional subspace by a projection matrix. Although, the effect of feature extraction is often worse than it in deep learning, it has always been a research hotspot because of its strong interpretability and particularly well on any type of hardware (CPU, GPU, DSP).  Therefore, it is an urgent need in traditional feature extraction to better extract discriminative features for downstream tasks.\par

In the field of deep learning, contrastive learning has attracted extensive scholarly attention as the primary method of self-supervised learning. Contrastive learning uses information of data to supervise itself by constructing positive and negative samples, which strives to learn more discriminative features. InfoNCE loss based on contrastive learning is proposed in contrastive predictive coding (CPC)\cite{1}.  CPC proves that minimizing the InfoNCE loss maximizes a lower bound on mutual information, which provides theoretical support for its advantages in extracting more discriminative features. Consequently, a large number of studies based on contrastive learning are proposed. Tian et al. proposed contrastive multiview coding (CMC)\cite{2} to process multi-view data. First, CMC constructs the same sample in any two views as positive samples and distinct samples as negative samples, and subsequently optimizes a neural network by minimizing the InfoNCE loss to maximize the similarity of the projected positive samples. Subsquently, Chen et al. propose a simple framework for contrastive learning (SimCLR)\cite{3} to process single-view data. First, it performs data enhancement to obtain different representations of samples, and then considers the representations of the same sample as positive samples and  the representations of distinct samples as negative samples. Finally, SimCLR optimizes the network by minimizing the InfoNCE loss, similar to CMC. In addition,  supervised contrastive learning (SupCon)\cite{4} has proved that contrastive learning also has great advantages in supervised learning. SupCon defines the samples in same class as positive samples and the samples in distinct classes as negative samples after data enhancement, and subsequently minimize the InfoNCE loss. Although these methods based on contrastive learning has good performance in both unsupervised and supervised learning, it also has some disadvantages. Firstly, the existing algorithms based on contrastive learning are produced in the field of deep learning and are not suitable for the traditional singal-view feature extraction problems. Secondly, even if we construct the InfoNCE loss according to these existing methods of defining positive and negative samples, like SimCLR and SupCon, and use it to perform  feature extraction, there are still some problems. For example, data enhancement will increase the running time of the algorithm. Moreover, the definitions of positive and negative samples ignore the structure information of data in unsupervised and supervised learning, which will lead to dispersion of samples in the same class and aggregation of samples in distinct classes, so that it hinders the extraction of more discriminative features. \par

Inspired by our prior research, we propose a feature extraction framework based on contrastive learning with adaptive positive and negative samples (CL-FEFA) that is suitable for unsupervised, supervised, and semi-supervised single-view feature extraction. The proposed framework, CL-FEFA, constructs adaptively the positive and negative samples from potential structural information based on the results of feature extraction, and the discriminative features are re extracted to according to infoNCE loss based on previous postive and negative samples. By leveraging the interactions between these two essential tasks, we are able to  construct more appropriate positive and negative samples, and extract more discriminative features. In addition, indicating matrix is introduced to unify the aspects of unsupervised, supervised, and semi-supervised feature extraction.
Furthermore, the effectiveness of the proposed framework is verified on four real-word datasets, including Yale, ORL, MINST, and CIFRA-10. \par
The main contributions of this study are as follows:
\begin{itemize}
	\item A feature extraction framework based on contrastive learning with adaptive positive and negative samples (CL-FEFA) is proposed from a new perspective that is suitable for unsupervised, supervised, and semi-supervised cases.
	\item CL-FEFA proposes a novel approach to construct adaptively more appropriate positive and negative samples in contrastive learning, and makes the model more robust. 
	\item It is proved that CL-FEFA actually maximizes the mutual information of similar samples in potential structures.
	\item The experiments on four real-word datasets show the advantages of the proposed framework.
\end{itemize}
\par 

The remainder of this article is organized as follows. The traditional feature extraction methods are briefly introduced in Section \uppercase\expandafter{\romannumeral2}. Subsequently, the development of the feature extraction framework (CL-FEFA) is discussed in Section \uppercase\expandafter{\romannumeral3}. In addition, the extensive experiments conducted on several real-world datasets are presented in Section \uppercase\expandafter{\romannumeral4}. Finally, the conclusions of the current study are detailed in Section \uppercase\expandafter{\romannumeral5}.

\section{Related Work}
In recent years, it has been witnessed that several important structures should be preserved in unsupervised, supervised, and semi-supervised feature extraction\cite{10}. Concretely, for unsupervised learning, locality preserving projections (LPP)\cite{5}, neighborhood preserving embedding (NPE)\cite{6}, sparsity preserving projections (SPP)\cite{7}, collaborative representation-based projections (CRP)\cite{8}, and low-rank preserving embedding (LRPE)\cite{9} are designed based on various graphs, respectively. Furthermore, supervised feature extraction methods obtain more discriminant information using sample labels in addition to preserving manifold structure. For example, local Fisher discriminant analysis (LFDA)\cite{11} combines the ideas of linear discriminant analysis (LDA)\cite{12} and LPP to locally construct the within-class scatter and between-class scatter, which achieve maximum preservation of the within-class and between-class local structures at the same time. Marginal Fisher analysis (MFA)\cite{10} considers the local structure within the class and constructs the local structure relationship between classes by accounting for the samples on the edges of various classes. As an improvement, multiple marginal Fisher analysis (MMFA)\cite{13} selects the nearest neighbor samples on all heterogeneous edges to construct the local relationship between classes. Sparsity preserving discriminant projections (SPDP)\cite{14} is proposed based on SPP to maintain the sparse reconstruction coefficients of within-class samples in the subspace. In addition, for semi-supervised feature exraction, Zhang et al. proposed semi-supervised LPP (SLPP), which preserves the manifold structure of labeled and unlabeled data, simultaneously. Huang et al. proposed a semi-supervised marginal Fisher analysis (SSMFA), which also preserves the manifold structure of labeled and unlabeled data, and it assigns discriminative weights to the edges of the different sample pairs. Liao et al. proposed a nonparameter framework, which is termed semi-supervised local discriminant analysis (SELD)\cite{15}. SELD aims to exploit the local neighbor information of unlabeled data while simultaneously preserving
the discriminant information of labeled data.\par 

However, the intrinsic structures designed by the above methods are unreliable and inaccurate when the redundant and noisy features are not removed. To address this, some adaptive structure preserving methods are proposed, which learn the structure information after feature extraction. Concretely, unsupervised feature extraction using a learned graph with clustering structure (LGCS), locality adaptive discriminant analysis (LADA), and semi-supervised adaptive local embedding learning (SALE) have been proposed in unsupervised, supervised, and semi-supervised cases, respectively.\par

Inspired by traditional methods, we propose a feature extraction framework based on contrastive learning with adaptive positive and negative samples (CL-FEFA). Concretely, compared with the previous models based on contrastive learning, CL-FEFA does not need data enhancement, and it constructs adaptively the positive and negative samples from the results of feature extraction, which makes it more appropriate and accurate.  Thereafter, the discriminative features are re extracted to according to InfoNCE loss based on previous positive and negative samples, which will make the intra-class samples more compact and the inter-class samples more dispersed. At the same time, using the potential structure information of subspace samples to dynamically construct positive and negative samples can make our framework more robust to noisy data. Compared with the traditional models, CL-FEFA is suitable for both unsupervised, supervised, and semi-supervised feature extraction, and it considers the mutual information between postive samples, that is, similar samples in potential structures.

\section{Methodology}
In this section, a feature extraction framework based on contrastive learning with adaptive positive and negative samples (CL-UFEF) is proposed for unsupervised, supervised, and semi-supervised feature extraction. 

Let us mathematically formulate the unsupervised, supervised, and semi-supervised feature extraction problem as follows. \par  
Feature extraction problem:  Given a training sample set $X=[x_1,x_2, . . .,x_n]\in{R^{D\times n}}$, where $n$ and $D$ are the number of samples and features, respectively. 
In the supervised case, labels for all samples are provided, and the label of $x_i, i=1,2,...,n$ is defined as $c_i$. In the semi-supervised case, label for a small number of samples are provided,  and the label of labeled sample $x_i$ is also defined as $c_i$. The purpose of feature extraction is to find a projection matrix $P\in{R^{D\times d}}$ to derive the low-dimensional embedding $Y = [y_1, y_2, . . ., y_n]\in{R^{d\times n}}$ for $X$ calculated by $Y = P^TX$, where $d\ll D$.\par 
For convenience, the symbols used in this study are summarized in Table \ref{Table1}.\par

\begin{table}[!ht]      
	\centering
	\caption{Notations and definitions.}
	\label{Table1}
	\begin{tabular}[t]{l l}
		\hline
		$X$ & Training sample set\\
		$Y$ & Set of training samples in a low-dimensional space\\
		$n$ & Number of training samples\\
		$D$ & Dimensionality of the samples in the original space\\
		$d$ & Dimensionality of embedding features\\
		$C$& The number of classes\\
		$H$ & Indicating matrix\\
		$S$ & Similarity matrix\\
		$c_i$ & Labels of sample $x_i$\\
		$P$ & Projection matrix\\
		$\sigma$ & Positive parameter\\
		$\lambda$ &Postive parameter\\
		$c$& Positive integer parameter\\
		k & Number of neighbors\\
		$NK(x_j)$ &The $k$ nearest neighbors of $x_j$\\
		$\nabla L(P)$ & Gradient of $L(P)$ with respect to $P$\\
		$T$ & Number of iterations\\
		\hline
	\end{tabular}
\end{table}	

\subsection{Framework of CL-FEFA}
In order to unify the aspects of unsupervised, supervised, and semi-supervised feature extraction, we first define an indicating matrix:
\begin{center}
	\begin{equation}
	H_{i,j}=\left \{
	\begin{aligned}
	&0&{\rm if}\, x_i\,{\rm and}\, x_j \,{\rm are\, labled}\,{\rm and} \,c_i\neq c_j;\\
	&1&{\rm otherwise}.
	\end{aligned}
	\right.
	\end{equation}
\end{center}
Thereafter,  we construct adaptively positive and negative samples by jointing indicating matrix $[H_{i,j}]_{n\times n}$ and the similarity matrix $[S_{i,j}]_{n\times n}$, where $S_{i,j}$ represents the similarity relationship of samples $x_i$ and $x_j$ in potential structure based
on the results of feature extraction. Contretely, $x_i$ and $x_j$ are defined as positive samples if $H_{i,j}S_{i,j}\neq 0$, and $x_i$ and $x_j$ are defined as negative samples if $H_{i,j}S_{i,j}=0$. Subsquently, the more discrimitive features are re extracted to according to InfoNCE loss based on postive and negative samples, and the projections of the positive samples $x_i$ and $x_j$ with larger $S_{i,j}$ should have greater similarity. Specifically, this optimization problem is defined as follows

\begin{center}
	\begin{equation}\label{CL-FEFA}
	\begin{aligned}
	&\min_{P,S}L=\sum_{i=1}^{n}\sum_{j=1}^{n}-H_{i,j}S_{i,j}log\frac{ f(y_i,y_j)}{\sum_{k=1}^{n}f(y_i,y_k)}+\gamma||S||_F^2\\
	&s.t.\; \forall i, S_{i}^T\bm{1}=1, 0\leqslant S_{i} \leqslant 1, rank(L_S)=n-c
	\end{aligned}
	\end{equation}
\end{center}
where

\begin{center}
	\begin{equation}
	f(y_i,y_j)=exp\left( \frac{y_i^Ty_j}{\|y_i\|\|y_j\|\sigma}\right) 
	\end{equation}
\end{center}
$\gamma$ and $\sigma$ are two positive parameters, $S_i$ represents $i$-th column vector of $S$, $\bm{1}$ is a n-dimensional column vector of all 1.  $rank(L_S)=n-c$ constraints the connected components of $S$ are exact $c$. In particular, in unsupervised and semi-supervised cases, $c$ is a positive integer parameter, and in supervised case, $c$ is the number of classes.\par 

\subsection{Relationship between CL-FEFA and Mutual Information}
For convenience, we make $W_{i,j} = H_{i,j}S_{i,j}, i,j=1,2,...,n$. Therefore, $x_j$ is a positive sample of $x_i$ iff $W_{i,j}\neq 0$, otherwise $x_j$ is a negative sample of $x_i$. Naturally, the probability that the sample in $X$ is a positive sample of $x_i$ is $p(W_{i,j}\neq 0|x_j, x_i)$, and the optimization problem (\ref{CL-FEFA}) is equivalent to
\begin{center}
	\begin{equation}\label{CL-FEFA01}
	\begin{aligned}
	&\min_{P,S}L=\sum_{i=1}^{n}\sum_{j=1}^{n}-W_{i,j}log\left[ p(W_{i,j}\neq 0|y_j, y_i)\right]
	+\gamma||S||_F^2\\
	&\;\;\;\;\;\;\;\;\;\;=\sum_{i=1}^{n}l_i++\gamma||S||_F^2\\
	&s.t.\; \forall i, S_{i}^T\bm{1}=1, 0\leqslant S_{i} \leqslant 1, rank(L_S)=n-c
	\end{aligned}
	\end{equation}
\end{center}
where $l_i=\sum_{j=1}^{n}-W_{i,j}log\left[ p(W_{i,j}\neq 0|y_j, y_i)\right]$.\par 
We define that the number of positive samples of $x_i$ is $n_i$, then the number of negative samples of $x_i$ is $n-n_i-1$. Therefore, there is two prior distribution $p(W_{i,j}\neq 0) = \frac{n_i}{n}$ and $p(W_{i,j}=0) = \frac{n-n_i-1}{n}$ when $i$ is fixed. According to the Bayesian formula, the following derivation is made
\begin{equation}
\begin{aligned}
p(W_{i,j}\neq 0|y_j, y_i)&=\frac{p(y_j, y_i|W_{i,j}\neq 0)p(W_{i,j}\neq 0)}{p(y_j, y_i|W_{i,j}\neq 0)p(W_{i,j}\neq 0)+p(y_j, y_i|W_{i,j}=0)p(W_{i,j}=0)}\\
&=\frac{p(y_j, y_i|W_{i,j}\neq 0)\frac{n_i}{n}}{p(y_j, y_i|W_{i,j}\neq 0)\frac{n_i}{n}+p(y_j, y_i|W_{i,j}=0)\frac{n-n_i-1}{n}}\\
&=\frac{n_ip(y_j, y_i|W_{i,j}\neq 0)}{n_ip(y_j, y_i|W_{i,j}\neq 0)+(n-n_i-1)p(y_j, y_i|W_{i,j}=0)}\\
&=\frac{n_ip(y_j, y_i)}{n_ip(y_j, y_i)+(n-n_i-1)p(y_j)p(y_i)}\\
\end{aligned}
\end{equation}
Further derivation, there is the following formula
\begin{equation}
\begin{aligned}
l_i&=\sum_{j=1}^{n}-W_{i,j}log\left[ p(W_{i,j}\neq 0|y_j, y_i)\right]\\
&=\sum_{j=1}^{n}-W_{i,j}log\left[ \frac{n_ip(y_j, y_i)}{n_ip(y_j, y_i)+(n-n_i-1)p(y_j)p(y_i)}\right]\\
&=\sum_{j=1}^{n}W_{i,j}log\left[ \frac{n_ip(y_j, y_i)+(n-n_i-1)p(y_j)p(y_i)}{n_ip(y_j, y_i)}\right]\\
&=\sum_{j=1}^{n}W_{i,j}log\left[ 1+\left( \frac{n-1}{n_i}-1\right) \frac{p(y_j)p(y_i)}{p(y_j, y_i)}\right]\\
&=\sum_{j=1}^{n}W_{i,j}log\left[ \frac{p(y_j,y_i)-p(y_j)p(y_i)}{p(y_j,y_i)}+ \frac{n-1}{n_i} \frac{p(y_j)p(y_i)}{p(y_j,y_i)}\right]\\
\end{aligned}
\end{equation}
Since $x_i$ and $x_j$ are positive samples, $y_i$ and $y_j$ are not independent, so $p(y_j,y_i)-p(y_j)p(y_i)>0$. Through the optimization problem (\ref{CL-FEFA01}), it can be seen that in the two potential tasks, the larger $W_{i,j}$ will lead to a larger $p(y_j,y_i)$, and the larger $p(y_j,y_i)$ will also lead to a larger $W_{i,j}$. In addition, $\sum_{j=1}^{n}W_{i,j}=\sum_{j=1}^{n}p(y_j,y_i)=1$, so $W_{i,j}\approx p(y_j,y_i)$. Therefore, we can get the following derivation
\begin{equation}
\begin{aligned}
l_i&=\sum_{j=1}^{n}p(y_j,y_i)log\left[ \frac{p(y_j,y_i)-p(y_j)p(y_i)}{p(y_j,y_i)}+ \frac{n-1}{n_i} \frac{p(y_j)p(y_i)}{p(y_j,y_i)}\right]\\
&\geqslant\sum_{j=1}^{n}p(y_j,y_i)log\left[ \frac{n-1}{n_i} \frac{p(y_j)p(y_i)}{p(y_j,y_i)}\right]\\
&= log\left[ \frac{n-1}{n_i}\right] -I(x_j,x_i)
\end{aligned}
\end{equation}
Therefore, we can get $-l_i\leqslant I(x_j,x_i)-log\left[ \frac{n-1}{n_i}\right]$, and minimizing $L$ in Eq. (\ref{CL-FEFA}) is equivalent to maximizing the mutual information of all positive samples, that is, similar samples in potential structures. \par

\subsection{Optimization Strategy}
Further, this optimization problem can be transformed as follows
\begin{center}
	\begin{equation}\label{CL-FEFA1}
	\begin{aligned}
	&\min_{P,S,F}L=\sum_{i=1}^{n}\sum_{j=1}^{n}-H_{i,j}S_{i,j}log\frac{ f(y_i,y_j)}{\sum_{k=1}^{n}f(y_i,y_k)}+\gamma||S||^2+2\lambda Tr\left( F^TL_SF\right) \\
	&s.t.\;\forall i, S_{i}^T\bm{1}=1, 0\leqslant S_{i} \leqslant 1, F\in R^{n\times c}, F^TF=I
	\end{aligned}
	\end{equation}
\end{center}
where $\lambda$ are a positive parameter, $L_S=D_S-\frac{S+S^T}{2}$ is called Laplacian matrix in graph theory, the degree matrix $D_S\in R^{n\times n}$ is defined as a diagonal matrix where the $i$-th diagonal element is $(S_{i,j} + S_{j,i})/2$.\par 
(1) When $P$ and $S$ are fixed, the optimization problem (\ref{CL-FEFA1}) becomes
\begin{center}
	\begin{equation}\label{F}
	\begin{aligned}
	&\min_{F}\;Tr\left( F^TL_SF\right) \\
	&s.t.\;F\in R^{n\times c}, F^TF=I
	\end{aligned}
	\end{equation}
\end{center}
The optimal solution $F$ to the problem (\ref{F}) is formed by the $c$ eigenvectors of $L_S$ corresponding to the $c$ smallest eigenvalues.\par 
(2) When $P$ and $F$ are fixed, the optimization problem (\ref{CL-FEFA1}) becomes
\begin{center}
	\begin{equation}\label{S}
	\begin{aligned}
	&\min_{S}\sum_{i=1}^{n}\sum_{j=1}^{n}-H_{i,j}S_{i,j}log\frac{ f(y_i,y_j)}{\sum_{k=1}^{n}f(y_i,y_k)}+\gamma||S||^2+2\lambda Tr(F^TL_SF)\\
	&s.t.\;\forall i, S_{i}^T\bm{1}=1, 0\leqslant S_{i} \leqslant 1
	\end{aligned}
	\end{equation}
\end{center}
The problem (\ref{S}) can be rewritten as
\begin{center}
	\begin{equation}\label{S1}
	\begin{aligned}
	&\min_{S}\sum_{i=1}^{n}\sum_{j=1}^{n}\left( -H_{i,j}S_{i,j}log\frac{ f(y_i,y_j)}{\sum_{k=1}^{n}f(y_i,y_k)}+\gamma S_{i,j}^2+\lambda ||f_i-f_j||^2\right) \\
	&s.t.\;\forall i, S_{i}^T\bm{1}=1, 0\leqslant S_{i} \leqslant 1
	\end{aligned}
	\end{equation}
\end{center}
Note that the problem (\ref{S1}) is independent between different $i$, so we can solve the following problem individually for each $i$:
\begin{center}
	\begin{equation}\label{S2}
	\begin{aligned}
	&\min_{S_i}\sum_{j=1}^{n}\left( -H_{i,j}S_{i,j}log\frac{ f(y_i,y_j)}{\sum_{k=1}^{n}f(y_i,y_k)}+\gamma S_{i,j}^2+\lambda ||f_i-f_j||^2S_{i,j}\right) \\
	&s.t.\;S_{i}^T\bm{1}=1, 0\leqslant S_{i} \leqslant 1
	\end{aligned}
	\end{equation}
\end{center}
Denote $d^y_{i,j}=-H_{i,j}log\frac{ f(y_i,y_j)}{\sum_{k=1}^{n}f(y_i,y_k)}$ and $d^F_{i,j}=||f_i-f_j||^2$, and $d_i\in R^{n\times 1}$ as a vector with the $j$-th elements as $d_{i,j}=d^y_{i,j}+d^f_{i,j}$, then the problem (\ref{S2}) can be written in vector form as
\begin{center}
	\begin{equation}\label{S3}
	\begin{aligned}
	&\min_{S_i}\left\|S_i+\frac{1}{2\gamma}d_i\right\| ^2\\
	&s.t.\; S_{i}^T\bm{1}=1, 0\leqslant S_{i} \leqslant 1
	\end{aligned}
	\end{equation}
\end{center}
The Lagrangian function of problem (\ref{S3}) is
\begin{center}
	\begin{equation}\label{S4}
	\mathcal{L}(S_i,\eta,\beta_i)=\frac{1}{2}\left\|S_i+\frac{1}{2\gamma}d_i\right\| ^2-\eta(S_i^T\bm{1}-1)-\beta_i^TS_i
	\end{equation}
\end{center}
where $\eta$ and $\beta_i \geqslant 0$ are the Lagrangian multipliers.\par 
According to the KKT condition, it can be verified that the optimal solution $S_i$ should be
\begin{center}
	\begin{equation}\label{S5}
	S_{i,j}=\left(-\frac{d_{i,j}}{2\gamma_i}+\eta\right)_+
	\end{equation}
\end{center}

Without loss of generality, suppose $d_{i,1}, d_{i,2}, ..., d_{i,n}$ are ordered
from small to large. If the optimal $S_i$ has only $k$
nonzero elements, then according to (\ref{S5}), we know $S_{i,k} >
0$ and $S_{i,k+1} = 0$. Therefore, we have
\begin{center}
	\begin{equation}\label{S6}
	\left\{
	\begin{aligned}
	&-\frac{d_{i,k}}{2\gamma_i}+\eta>0\\
	&-\frac{d_{i,k+1}}{2\gamma_i}+\eta\leqslant 0\\
	\end{aligned}\right.
	\end{equation}
\end{center}
According to (\ref{S5}) and the constraint $S_i^T\bm{1} = 1$, we have
\begin{center}
	\begin{equation}\label{S7}
	\begin{aligned}
	&\sum_{j=1}^{k}(-\frac{d_{i,j}}{2\gamma_i}+\eta)=1\\
	&\Rightarrow\eta=\frac{1}{k}+\frac{1}{2k\gamma_i}\sum_{j=1}^{k}d_{i,j}\\
	\end{aligned}
	\end{equation}
\end{center}
So we have the following inequality for $\gamma_i$ according to (\ref{S6}) and (\ref{S7}):
\begin{center}
	\begin{equation}\label{S8}
	\frac{k}{2}d_{i,k}-\frac{1}{2}\sum_{j=1}^{k}d_{i,j}<\gamma_i<\frac{k}{2}d_{i,k+1}-\frac{1}{2}\sum_{j=1}^{k}d_{i,j}
	\end{equation}
\end{center}
Therefore, in order to obtain an optimal solution $S_i$ that has exact $k$ nonzero values, we could set $\gamma_i$ to be
\begin{center}
	\begin{equation}\label{S9}
	\gamma_i=\frac{k}{2}d_{i,k+1}-\frac{1}{2}\sum_{j=1}^{k}d_{i,j}
	\end{equation}
\end{center}
The overall $\gamma$ could be set to the mean of $\gamma_1, \gamma_2, ..., \gamma_n$. That is, we could set the $\gamma$ to be
\begin{center}
	\begin{equation}\label{S10}
	\gamma=\frac{1}{n}\sum_{i=1}^{n}\left(\frac{k}{2}d_{i,k+1}-\frac{1}{2}\sum_{j=1}^{k}d_{i,j}\right)
	\end{equation}
\end{center}
The number of neighbors $k$ is much easier to tune than the regularization parameter $\gamma$ since $k$ is an integer and has explicit meaning.

(3) When $S$ and $F$ are fixed, the optimization problem (\ref{CL-FEFA1}) becomes
	\begin{equation}\label{P}
	\min_{P}\;{L(P)}=\sum_{i=1}^{n}\sum_{j=1}^{n}-H_{i,j}S_{i,j}log\frac{ f(y_i,y_j)}{\sum_{k=1}^{n}f(y_i,y_k)}
	\end{equation}
The problem (\ref{P}) is solved by using the Adam optimizer. Adam is an advancement on the random gradient descent method and can rapidly yield accurate results. This method calculates the adaptive learning rate of various parameters based on the budget of the first and second moments of the gradient. The parameters $\alpha$, $\beta_1$, $\beta_2$, and $\epsilon$ represent the learning rate, the exponential decay rate of the first- and second-order moment estimation, and the parameter to prevent division by zero in the implementation, respectively. In addition, the gradient of the loss function with respect to the projection matrix $P$ is obtained from (\ref{grad}).
\begin{center}
	\begin{equation}\label{grad} 
	\begin{aligned}
	&\nabla L(P)=\\
	&\sum_{i=1}^{n} \sum_{j=1}^{n}\{ -H_{i,j}S_{i,j}\frac{\sum_{k=1}^{n}f(y_i,y_k)}{f(y_i,y_j)}\cdot\\
	&\frac{[\nabla f(y_i,y_j)\cdot\sum_{k=1}^{n}f(y_i,y_k)-\sum_{k=1}^{n}\nabla f(y_i,y_k)\cdot f(y_i,y_j)] }{\left[ \sum_{k=1}^{n}f(y_i,y_j)\right] ^2}\}\\  
	\end{aligned}
	\end{equation}
\end{center}
where
\begin{center}
	\begin{equation}
	\begin{aligned}
	&\nabla f(y_i,y_j)=\\
	&\{(x_i{x_j}^T+x_j{x_i}^T)P\cdot\|P^Tx_i\|\|P^Tx_j\|\sigma\\
	&-[({x_i}^TPP^Tx_i)^{-\frac{1}{2}}\cdot x_i{x_i}^TP\cdot\|P^Tx_j\|\sigma\\
	&+({x_j}^TPP^Tx_j)^{-\frac{1}{2}}\cdot x_j{x_j}^TP\cdot\|P^Tx_i\|\sigma]\cdot {x_i}^TPP^Tx_j\}/\\
	&{(\|P^Tx_i\|\|P^Tx_j\|\sigma)}^2
	\end{aligned}
	\end{equation}
\end{center}

So far, the optimization steps of three variables have been described at all. we summarize our optimization algorithm in the following Algorithm 1. 
The convergent condition used in our experiments is set as $\left| L(P_t) - L(P_{t+1})\right| \leqslant 10^{-3}$ and $\left| L(P_T) - L(P_{T+1})\right| \leqslant 10^{-3}$.
\begin{algorithm}[!h]\label{Algorithm1}
	\caption{CL-FEFA} 
	{\bf Input:} 
	
	Data matrix: $X\in R^{D\times n}$, $d$, $\alpha,\beta_1,\beta_2,\epsilon,P_0,S_0$.\\
	$T= 0$ (Initialize number of iterations)\\
	\hspace*{0.02in} {\bf Output:} 
	Projection matrix $P$
	\begin{algorithmic}
		\WHILE{$P_T$ not converged} 
		\STATE
		Update $F$ by (\ref{F})\\
		Update $S$ by (\ref{S5})\\
		$m_0=0 $(Initialize $1^{st}$ moment vector)\\
		$v_0=0 $(Initialize $2^{nd}$ moment vector)\\
		$t= 0$ (Initialize descent times)\\
		\WHILE {$P_t$ not converged}
		\STATE 
		$g_t=\bigtriangledown L(P_{t-1})$ is calculated using (\ref{gradient}) (Obtain gradients with respect to the stochastic objective at time step $t$)\\
		$m_t=\beta_1\cdot m_{t-1}+(1-\beta_1)\cdot g_t$ (Update biased first-moment estimate)\\
		$v_t=\beta_2\cdot v_{t-1}+(1-\beta_2)\cdot g_t^2$ (Update biased second raw-moment estimate)\\
		$\hat{m}_t=m_t/(1-\beta_1^t)$ (Compute bias-corrected first-moment estimate)\\
		$\hat{v}_t=v_t/(1-\beta_2^t)$ (Compute bias-corrected second raw-moment estimate)\\
		$P_t=P_{t-1}-\alpha\cdot\hat{m}_t/(\sqrt{\hat{v}_t}+\epsilon)$\\
		$t=t+1$\\
		\ENDWHILE\\
		$P_T=P_t$\\
		$T=T+1$\\
		\ENDWHILE
		\RETURN P
	\end{algorithmic}
\end{algorithm}
\subsection{Complexity analysis}
The main computational complexity of each cycle in Algorithm 1 is the derivation of the loss function in the first step of Adam optimizer, which is $O(n^2(D^2d+Dd+D^2))$. Assuming that the Algorithm 1 performs a total of $M$ iterations and $m$ cycles in Adam optimizer of each iteration when converging, the main computational complexity is $O(Mmn^2(D^2d+Dd+D^2))$.

\section{Experimental results}
To demonstrate the effectiveness of the proposed CL-FEFA, four datasets are utilized in our experiments. The comparison methods in unsupervised, supervised, and semi-supervised cases are as follows:\par 
In unsupervised case, the performance of u-CL against LPP, FLPP, LAPP, SimCLR, where: LPP is a typical unsupervised method which aims at preserving local neighbor information of the samples; FLPP is a new unsupervised method which aims at preseving local neighbor information of the samples; LAPP is a new unsupervised method which aims at preseving local neighbor information and is insensitive to noise of samples; SimCLR is a new unsupervised deep learning method based on constrastive learning, defines the same sample as positive pairs and the distinct samples as negative pairs through data enhancement. Note that in order to use SimCLR for feature extraction, data enhancement is performed by rotating each picture in the data 90 degrees counterclockwise, and then its loss function is used to obtain projection matrix $P$. \par 
In supervised case, the performance of s-CL against LDA, FDLPP, LADA, SupCon, where: LDA is a typical supervised method which aims at minimizing the within-class scatter and maximizing the between-class scatter without local preservation; FDLPP  is a new supervised method which aims at minimizing the within-class scatter and maximizing the between-class scatter with local preservation; LADA is a new supervised method which aims at minimizing the within-class scatter and maximizing the between-class scatter with adaptive local preservation; SupCon is new supervised deep learning method based on constrastive learning, defines the samples of same class as positive pairs and the samples of distinct classes as negative pairs after data enhancement. Moreover, data enhancement is performed just like SimCLR.\par 
In semi-supervised case, the performance of semi-CL against SLPP, SELD, SSMFA, SALWE, where: SLPP is a typical semi-supervised method which aims at preserving the manifold structure of labeled and unlabeled data, simultaneously. SELD is a typical semi-supervised method which aims to exploit the local neighbor information of unlabeled data while simultaneously preserving the discriminant information of labeled data. SSMFA is a typical semi-method which aims at preserving the manifold structure of labeled and unlabeled data, and it assigns discriminative weights to the edges of the different sample pairs. SALWE is a new semi-supervised method which aims at adaptive preserving the manifold structure of labeled and unlabeled data.\par 

\subsection{Dataset description}

Yale dataset: The dataset is created by Yale University Computer Vision and Control Center, containing data of $15$ individuals, wherein each person has 11 frontal images ($64\times64$ pixels in size) captured under various lighting conditions. The images are edited to $50\times40$ pixels with 256 Gy levels per pixel.\par 
ORL dataset: The ORL dataset contains 400 images of 40 different people. Each image of the same person is captured in different time, light, facial expressions (open eyes/close eyes, smile/no smile) and facial details (glasses/no glasses). All images are are sized $50\times 40$ pixels with 256-level gray scale.\par 
MNIST dataset: This dataset contains 70,000 samples of $0--9$ digital images with a size of $28\times28$. We randomly select 2000 images as experimental data, uniformly rescale all the images to a size of $16\times16$, and use a feature vector of 256-level grayscale pixel values  to represent each image.\par 
CIFAR-10 dataset:  This dataset contains 60,000 samples of  $32\times32$ color images in 10 categories (airplane, automobile, bird, cat, deer, dog, frog, horse, ship, truck). Each image is represented by a 3072 dimensional feature vector, in which the first 1024 features represent the red channel value, the next 1024 features represent the green channel value, and the last 1024 features represent the blue channel value.\par

\subsection{Experimental Setups}

\begin{table*}[!ht]      
	\centering
	\caption{Description of datasets.}
	\label{Table2}
	\begin{tabular}[t]{c c c c c }
		\hline
		Datasets  &Samples & Dimensions & Classes  & Training samples\\
		\hline
		Yale&165&2000&15 & 60\\
		ORL&400&2000&40 & 60\\
		MNIST&2000&256&10 & 60\\
		CIFAR-10&60000&3072&10 &60\\
		\hline
	\end{tabular}
\end{table*}	

To fully assess the effectiveness of our proposed CL-FEFA, we show that our methods perform well on classification task. 
The k-nearest neighbor classifier (k = 1) is used in the experiment. Moreover, four samples of each class from Yale and ORL,  six samples of each class from MNIST and CIFAR-10 datasets, are randomly selected for training, and the remaining data are used for testing, and the details are listed in Table \ref{Table2}. All processes are repeated five times, and the final evaluation criteria constitute the average recognition accuracy and average recall rate of five repeated experiments. The calculation methods of recognition accuracy and recall rate are shown in (\ref{accuracy}) and (\ref{recall}). The experiments are implemented using MATLAB R2018a on a computer with an Intel Core i5-9400 2.90 GHz CPU and Windows 10 operating system.
\begin{equation}\label{accuracy}
{\rm Classification\;Accuracy}  =\frac{\sum_{c=1}^{C}T_c}{n}
\end{equation}

\begin{equation}\label{recall}
{\rm Recall\;Rate}  =\sum_{c=1}^{C}\frac{T_c}{n_c}/C
\end{equation}
where $T_c, c=1,...,C$ is the count of true samples in $c$th class, $n_c, c=1,...,C$ is the count of forecasting samples in $c$th class.

\subsection{Parameters Setting}
The performance of various feature extraction methods is evaluated by setting certain parameters in advance. First, the more appropriate default parameters for testing machine learning problems in Adam optimizer comprise $\alpha = 0.001$, $\beta_1 = 0.9$, $\beta_2 = 0.999$, and $\epsilon = 10^{-8}$. Thereafter, for all comparative algorithms, the search range of k is set to $\{2,6,10\}$, whereas the range of $\sigma$ for u-CL, s-CL, semi-CL, SimCLR and SupCon are set as $\{-2,-1,0,1,2,3\}\times 1e$. In addition, the parameter $\lambda$ is set as $\{-4,-2,0,2,4\}\times 1e$ for u-CL and semi-CL. In the supervised case, there is $H_{i,j}S_{i,j}$ = 0 for all heterogeneous samples $x_i$ and $x_j$, so in s-CL, we make $\lambda = 0$. At the same time, in order to shorten the running time of the algorithm, for u-CL and semi-CL, let the parameter $c$ be the total number of classes of each dataset, respectively.

\subsection{Results Analysis}

\begin{table*}[!ht]      
	\centering
	\caption{Major comparative items and motivations.}
	\label{Table3}
	\resizebox{\textwidth}{!}{
		\begin{tabular}[t]{l l}
			\hline
			Items &Motivations\\
			\hline
			\textbf{1. u-CL vs. unsupervised traditional methods}&to indicate that CL-FEFA is superior than traditional unsupervised methods\\
			\textbf{2. s-CL and supervised traditional methods}& to indicate that CL-FEFA is superior than traditional supervised methods\\
			\textbf{3. semi-CL vs. semi-supervised traditional methods} & to indicate that CL-FEFA is superior than semi-supervised traditional methods\\
			\hline
		\end{tabular}}
	\end{table*}	
	
	The superior performance of CL-FEFA is demonstrated by comparing the experimental results of all the above-mentioned methods. The major comparative items and motivations are summarized in Table \ref{Table3}.\par 
	First, we report the maximum mean classification accuracy (contains standard) and maximum mean recall rate (contains standard) deviation under optimal feature extraction on Yale, ORL, MNIST, and CIFRA-10 datasets for unsupervised, supervised, and semi-supervised case in Table \ref{Table4}, \ref{Table5}, and \ref{Table6}, where “Mean” represents the average of the four datasets. The most appropriate results for each dataset are marked in bold. In addition, the mean classification accuracy of all the methods under various reduced dimensions on each dataset is presented in
	Figure \ref{fig2}, \ref{fig3}, and \ref{fig4}. Based on the experimental results, the following observations are made.\par
	
	\noindent\textbf{Item 1. u-CL vs. unsupervised traditional methods}\par
	
	As can be observed from Table \ref{Table3}, the maximum mean classification accuracy of u-CL is higher than all comparison methods on all datasets, with an average of 9.55\%, 3.90\%, 5.34\%, and 4.45\% higher than those of LPP, FLPP, LAPP, and SimCLR, respectively. Moreover, the maximum mean recall rate of u-CL is higher than all comparison for all datasets, with an average of 8.06\%, 2.43\%, 3.75\%, and 3.20\% higher than
	those of LPP, FLPP, LAPP, and SimCLR respectively. \par
	
	\noindent\textbf{Item 2. s-CL vs. supervised traditional methods}\par
	
	As can be observed from Table \ref{Table4}, the maximum mean classification accuracy of s-CL is higher than all comparison methods on all datasets, with an average of 7.09\%, 5.08\%, 6.22\%, and 4.99\% higher than those of LDA, FDLPP, LADA, and SupCon, respectively. Moreover, the maximum mean recall rate of s-CL is higher than all comparison for all datasets, with an average of 7.27\%, 6.98\%, 5.33\%, and 5.87\% higher than hose of LDA, FDLPP, LADA, and SupCon, respectively. \par
	
	\noindent\textbf{Item 3. semi-CL vs. semi-supervised traditional methods}\par
	
	As can be observed from Table \ref{Table5}, the maximum mean classification accuracy of semi-CL is higher than all comparison methods on all datasets, with an average of 9.27\%, 7.81\%, 6.34\%, and 6.23\% higher than those of SLPP, SELD, SSMFA, and SALWE, respectively. Moreover, the maximum mean recall rate of semi-CL is higher than all comparison for all datasets, with an average of 14.41\%, 5.33\%, 10.99\%, and 4.11\% higher than those of SLPP, SELD, SSMFA, and SALWE, respectively. \par 
	\begin{table*}[!ht]      
		\centering
		\caption{Experimental results of unsupervised methods on optimal dimensions.}
		\label{Table4}
		\begin{tabular}{c c c c c c}
			\hline
			Datasets& LPP & FLPP &LAPP & SimCLR & u-CL \\
			\hline
			\multicolumn{6}{c}{ Classification Accuracy}\\
			\hline
			Yale    &$73.00\pm4.31$&$76.00\pm3.25$&$74.00\pm5.08$&$77.00\pm3.98$&\bm{$81.33\pm3.61$}\\
			ORL&$85.00\pm2.30$&$92.50\pm4.35$&$94.75\pm1.37$&$88.63\pm2.81$&\bm{$94.88\pm2.55$}
			\\
			MNIST&$72.00\pm5.70$&$83.33\pm4.86$&$74.00\pm2.24$&$84.00\pm5.08$&\bm{$85.33\pm4.15$}\\
			CIFRA-10&$37.41\pm5.14$&$38.15\pm9.85$&$41.48\pm4.65$&$38.15\pm6.36$&\bm{$44.07\pm3.31$}\\
			Mean&$66.85\pm4.36$&$72.50\pm5.58$&$71.06\pm3.34$&$71.95\pm4.56$&\bm{$76.40\pm3.41$}\\
			\hline
			\multicolumn{6}{c}{Recall Rate}\\
			\hline
			Yale    &$80.74\pm4.96$&$83.90\pm1.59$&$80.76\pm3.01$&$85.26\pm2.01$&\bm{$87.65\pm3.51$}\\
			ORL&$86.05\pm4.14$&$93.86\pm2.30$&$95.72\pm1.17$&$90.66\pm2.35$&\bm{$95.90\pm2.08$}
			\\
			MNIST&$76.35\pm5.58$&$84.63\pm4.67$&$79.38\pm1.45$&$85.67\pm4.76$&\bm{$86.65\pm4.01$}\\
			CIFRA-10&$42.96\pm7.01$&$46.24\pm4.27$&$47.49\pm5.57$&$43.97\pm6.66$&\bm{$48.14\pm5.67$}\\	
			Mean&$71.53\pm5.42$&$77.16\pm3.21$&$75.84\pm2.80$&$76.39\pm3.95$&\bm{$79.59\pm3.82$}\\
			\hline		
		\end{tabular}
	\end{table*}

	\begin{table*}[!ht]      
		\centering
		\caption{Experimental results of supervised methods on optimal dimensions.}
		\label{Table5}
		\begin{tabular}{c c c c c c}
			\hline
			Datasets& LDA & FDLPP &LADA & SupCon & s-CL\\
			\hline
			\multicolumn{6}{c}{Classification Accuracy}\\
			\hline
			Yale    &$78.00\pm4.62$&$80.67\pm3.25$&$76.67\pm3.54$&$76.67\pm1.18$&\bm{$84.67\pm1.83$}\\
			ORL    &$89.62\pm2.75$&$94.12\pm2.88$&$91.13\pm3.96$&$94.25\pm2.88$&\bm{$95.38\pm2.36$}\\
			MNIST&$83.00\pm2.74$&$83.33\pm6.12$&$84.67\pm5.94$&$85.00\pm6.35$&\bm{$91.00\pm5.08$}\\
			CIFRA-10&$35.78\pm4.06$&$36.30\pm2.11$&$37.41\pm0.83$&$38.89\pm4.14$&\bm{$43.70\pm5.94$}\\
			Mean&$71.60\pm3.54$&$73.61\pm3.59$&$72.47\pm3.57$&$73.70\pm3.64$&\bm{$78.69\pm3.80$}\\
			\hline
			\multicolumn{6}{c}{Recall Rate}\\
			\hline
			Yale    &$84.50\pm3.15$&$80.52\pm3.96$&$84.63\pm3.07$&$84.56\pm2.59$&\bm{$89.77\pm2.70$}\\
			ORL    &$91.99\pm2.27$&$95.29\pm2.36$&$92.92\pm3.35$&$90.90\pm1.61$&\bm{$96.25\pm2.06$}\\
			MNIST&$84.64\pm6.57$&$84.75\pm3.94$&$86.72\pm8.00$&$86.80\pm6.01$&\bm{$92.21\pm2.74$}\\
			CIFRA-10&$39.48\pm3.27$&$41.19\pm7.55$&$44.08\pm3.04$&$43.95\pm5.96$&\bm{$51.44\pm6.91$}\\
			Mean&$75.15\pm3.82$&$75.44\pm4.45$&$77.09\pm4.37$&$76.55\pm4.04$&\bm{$82.42\pm3.60$}\\
			\hline
		\end{tabular}
	\end{table*}

	\begin{table*}[!ht]      
		\centering
		\caption{Experimental results of semi-supervised methods on optimal dimensions.}
		\label{Table6}
		\begin{tabular}{c c c c c c}
			\hline
			Datasets& SLPP & SELD &SSMFA & SALWE & semi-CL \\
			\hline
			\multicolumn{6}{c}{Classification Accuracy}\\
			\hline
			Yale    &$83.00\pm2.74$&$84.00\pm3.46$&$84.33\pm4.50$&$83.67\pm4.31$&\bm{$93.71\pm2.48$}\\
			ORL&$92.13\pm2.28$&$95.00\pm2.69$&$94.12\pm1.37$&$95.12\pm1.62$&\bm{$96.08\pm1.30$}\\
			MNIST&$70.33\pm3.42$&$72.33\pm5.96$&$74.67\pm1.39$&$77.00\pm8.53$&\bm{$85.00\pm2.64$}\\
			CIFRA-10&$38.52\pm7.57$&$38.52\pm5.62$&$42.59\pm2.93$&$40.37\pm5.62$&\bm{$46.30\pm3.70$}\\
			Mean&$71.00\pm4.00$&$72.46\pm4.43$&$73.93\pm2.55$&$74.04\pm5.02$&\bm{$80.27\pm2.53$}\\
			\hline
			\multicolumn{6}{c}{Recall Rate}\\
			\hline
			Yale    &$81.79\pm2.93$&$88.97\pm1.67$&$83.21\pm4.82$&$88.78\pm2.90$&\bm{$95.02\pm2.34$}\\
			ORL    &$91.92\pm2.34$&$96.12\pm2.04$&$93.97\pm1.40$&$96.28\pm0.88$&\bm{$97.00\pm0.47$}\\
			MNIST
			&$67.04\pm3.80$&$77.75\pm3.79$&$71.85\pm1.55$&$79.50\pm8.32$&\bm{$86.36\pm3.58$}\\
			CIFRA-10
			&$30.83\pm8.51$&$44.22\pm6.31$&$35.42\pm3.29$&$47.38\pm8.03$&\bm{$50.02\pm1.89$}\\
			Mean&$67.90\pm4.40$&$76.77\pm3.45$&$71.11\pm2.77$&$77.99\pm5.03$&\bm{$82.10\pm2.07$}\\
			\hline
		\end{tabular}
	\end{table*}

	From the above experimental results, we can know that our proposed framework CL-FEFA shows obvious advantages in unsupervised, supervised, and semi-supervised feature extraction, whether the dataset is added with noise or not. In particular, compared with SimCLR and SupCon, the advantages of our framework prove that the method of adaptively constructing positive and negative samples in contrastive learning is more conducive to the traditional feature extraction problem.

\bibliographystyle{unsrtnat}
\bibliography{main}  

\begin{thebibliography}{15}
\providecommand{\natexlab}[1]{#1}
\providecommand{\url}[1]{\texttt{#1}}
\expandafter\ifx\csname urlstyle\endcsname\relax
  \providecommand{\doi}[1]{doi: #1}\else
  \providecommand{\doi}{doi: \begingroup \urlstyle{rm}\Url}\fi

\bibitem[van~den Oord et~al.(2018)van~den Oord, Li, and Vinyals]{1}
A{\"{a}}ron van~den Oord, Yazhe Li, and Oriol Vinyals.
\newblock Representation learning with contrastive predictive coding.
\newblock \emph{CoRR}, abs/1807.03748, 2018.

\bibitem[Tian et~al.(2020)Tian, Krishnan, and Isola]{2}
Yonglong Tian, Dilip Krishnan, and Phillip Isola.
\newblock Contrastive multiview coding.
\newblock In \emph{{ECCV} {(11)}}, volume 12356 of \emph{Lecture Notes in
  Computer Science}, pages 776--794. Springer, 2020.

\bibitem[Chen et~al.(2020)Chen, Kornblith, Norouzi, and Hinton]{3}
Ting Chen, Simon Kornblith, Mohammad Norouzi, and Geoffrey~E. Hinton.
\newblock A simple framework for contrastive learning of visual
  representations.
\newblock In \emph{{ICML}}, volume 119 of \emph{Proceedings of Machine Learning
  Research}, pages 1597--1607. {PMLR}, 2020.

\bibitem[Khosla et~al.(2020)Khosla, Teterwak, Wang, Sarna, Tian, Isola,
  Maschinot, Liu, and Krishnan]{4}
Prannay Khosla, Piotr Teterwak, Chen Wang, Aaron Sarna, Yonglong Tian, Phillip
  Isola, Aaron Maschinot, Ce~Liu, and Dilip Krishnan.
\newblock Supervised contrastive learning.
\newblock In \emph{NeurIPS}, 2020.

\bibitem[Yan et~al.({2007})Yan, Xu, Zhang, Zhang, Yang, and Lin]{10}
Shuicheng Yan, Dong Xu, Benyu Zhang, Hong-Jiang Zhang, Qiang Yang, and Stephen
  Lin.
\newblock {Graph embedding and extensions: A general framework for
  dimensionality reduction}.
\newblock \emph{{IEEE TRANSACTIONS ON PATTERN ANALYSIS AND MACHINE
  INTELLIGENCE}}, {29}\penalty0 ({1}):\penalty0 {40--51}, {JAN} {2007}.
\newblock ISSN {0162-8828}.
\newblock \doi{{10.1109/TPAMI.2007.250598}}.

\bibitem[He(2003)]{5}
X~He.
\newblock Locality preserving projections.
\newblock \emph{Advances in Neural Information Processing Systems}, 16\penalty0
  (1):\penalty0 186--197, 2003.

\bibitem[He et~al.(2005)He, Cai, and Yan]{6}
Xiaofei He, Deng Cai, and Shuicheng Yan.
\newblock Neighborhood preserving embedding.
\newblock volume~2, pages 1208-- 1213 Vol. 2, 11 2005.
\newblock ISBN 0-7695-2334-X.
\newblock \doi{10.1109/ICCV.2005.167}.

\bibitem[Qiao et~al.({2010})Qiao, Chen, and Tan]{7}
Lishan Qiao, Songcan Chen, and Xiaoyang Tan.
\newblock {Sparsity preserving projections with applications to face
  recognition}.
\newblock \emph{{PATTERN RECOGNITION}}, {43}\penalty0 ({1}):\penalty0
  {331--341}, {JAN} {2010}.
\newblock ISSN {0031-3203}.
\newblock \doi{{10.1016/j.patcog.2009.05.005}}.

\bibitem[Yang et~al.({2015})Yang, Wang, and Sun]{8}
Wankou Yang, Zhenyu Wang, and Changyin Sun.
\newblock {A collaborative representation based projections method for feature
  extraction}.
\newblock \emph{{PATTERN RECOGNITION}}, {48}\penalty0 ({1}):\penalty0 {20--27},
  {JAN} {2015}.
\newblock ISSN {0031-3203}.
\newblock \doi{{10.1016/j.patcog.2014.07.009}}.

\bibitem[Zhang et~al.({2017})Zhang, Xiang, and Yang]{9}
Yupei Zhang, Ming Xiang, and Bo~Yang.
\newblock {Low-rank preserving embedding}.
\newblock \emph{{PATTERN RECOGNITION}}, {70}:\penalty0 {112--125}, {OCT}
  {2017}.
\newblock ISSN {0031-3203}.
\newblock \doi{{10.1016/j.patcog.2017.05.003}}.

\bibitem[Sugiyama({2007})]{11}
Masashi Sugiyama.
\newblock {Dimensionality reduction of multimodal labeled data by local fisher
  discriminant analysis}.
\newblock \emph{{JOURNAL OF MACHINE LEARNING RESEARCH}}, {8}:\penalty0
  {1027--1061}, {MAY} {2007}.
\newblock ISSN {1532-4435}.

\bibitem[Martinez and Kak({2001})]{12}
AM~Martinez and AC~Kak.
\newblock {PCA versus LDA}.
\newblock \emph{{IEEE TRANSACTIONS ON PATTERN ANALYSIS AND MACHINE
  INTELLIGENCE}}, {23}\penalty0 ({2}):\penalty0 {228--233}, {FEB} {2001}.
\newblock ISSN {0162-8828}.
\newblock \doi{{10.1109/34.908974}}.

\bibitem[Huang et~al.({2019})Huang, Zhu, Zhou, and Peng]{13}
Zhenyu Huang, Hongyuan Zhu, Joey~Tianyi Zhou, and Xi~Peng.
\newblock {Multiple Marginal Fisher Analysis}.
\newblock \emph{{IEEE TRANSACTIONS ON INDUSTRIAL ELECTRONICS}}, {66}\penalty0
  ({12}):\penalty0 {9798--9807}, {DEC} {2019}.
\newblock ISSN {0278-0046}.
\newblock \doi{{10.1109/TIE.2018.2870413}}.

\bibitem[Ren et~al.({2016})Ren, Wang, Chen, and Zhao]{14}
Yingchun Ren, Zhicheng Wang, Yufei Chen, and Weidong Zhao.
\newblock {Sparsity Preserving Discriminant Projections with Applications to
  Face Recognition}.
\newblock \emph{{MATHEMATICAL PROBLEMS IN ENGINEERING}}, {2016}, {2016}.
\newblock ISSN {1024-123X}.
\newblock \doi{{10.1155/2016/5269236}}.

\bibitem[Liao et~al.(2013)Liao, Pizurica, Scheunders, Philips, and Pi]{15}
Wenzhi Liao, Aleksandra Pizurica, Paul Scheunders, Wilfried Philips, and Youguo
  Pi.
\newblock Semisupervised local discriminant analysis for feature extraction in
  hyperspectral images.
\newblock \emph{{IEEE} Trans. Geosci. Remote. Sens.}, 51\penalty0 (1):\penalty0
  184--198, 2013.

\end{thebibliography}






\end{document}